\documentclass[10pt,twocolumn,letterpaper]{article}

\usepackage{cvpr}              

\usepackage{graphicx}
\usepackage{amsmath}
\usepackage{amssymb}
\usepackage{booktabs}
\usepackage{url}
\usepackage{makecell, multirow}
\usepackage{algorithm}
\usepackage{algorithmic}

\usepackage[pagebackref,breaklinks,colorlinks]{hyperref}

\usepackage[capitalize]{cleveref}
\crefname{section}{Sec.}{Secs.}
\Crefname{section}{Section}{Sections}
\Crefname{table}{Table}{Tables}
\crefname{table}{Tab.}{Tabs.}

\def\cvprPaperID{8718} 
\def\confName{CVPR}
\def\confYear{2022}

\begin{document}

\title{RGP: Neural Network Pruning through Its Regular Graph Structure}

\author{
Zhuangzhi Chen, 
Jingyang Xiang, 
Yao Lu, 
Qi Xuan, 
Xiaoniu Yang\\
Zhejiang University of Technology\\
{\tt\small zzch@zjut.edu.cn, xiangxiangjingyang@gmail.com, yaolu.zjut@gmail.com}\\ {\tt\small xuanqi@zjut.edu.cn, yxn2117@126.com}
}
\maketitle

\begin{abstract}
  Lightweight model design has become an important direction in the application of deep learning technology, pruning is an effective mean to achieve a large reduction in model parameters and FLOPs. The existing neural network pruning methods mostly start from the importance of parameters, and design parameter evaluation metrics to perform parameter pruning iteratively. These methods are not studied from the perspective of model topology, may be effective but not efficient, and requires completely different pruning for different datasets. In this paper, we study the graph structure of the neural network, and propose regular graph based pruning (RGP) to perform a one-shot neural network pruning. We generate a regular graph, set the node degree value of the graph to meet the pruning ratio, and reduce the average shortest path length of the graph by swapping the edges to obtain the optimal edge distribution. Finally, the obtained graph is mapped into a neural network structure to realize pruning. Experiments show that the average shortest path length of the graph is negatively correlated with the classification accuracy of the corresponding neural network, and the proposed RGP shows a strong precision retention capability with extremely high parameter reduction (more than 90\%) and FLOPs reduction (more than 90\%).
\end{abstract}

\section{Introductions}
The architecture of deep neural network (DNN) is one of the most important factors affecting the performance of the model. DNNs with good performance are often carefully designed by experienced researchers. For instance, AlexNet~\cite{krizhevsky2012imagenet}, GoogleNet (Inception)~\cite{szegedy2015going}, VGG16~\cite{simonyan2014very}, ResNet~\cite{he2016deep} and DenseNet~\cite{huang2017densely}, whose appropriate structure designs enable the models to have powerful feature extraction and pattern recognition capabilities.

However, even a well-designed DNN has proved to have a large amount of parameter redundancy, which is called over-parameterization of DNN~\cite{NIPS2014_ea8fcd92,NIPS2014_2afe4567}. Users can remove those redundant parameters without affecting too much accuracy of the model. Such process is called the pruning of the neural network. The pruned neural network can usually obtain much faster inference speed than the original model, which has high significance in actual deployment when the efficiency of the model is critical. The pruning of neural network can usually be seen as a three-step pipeline: training the original model, parameter pruning, and fine-tuning the pruned model. Thus most of these network pruning methods are data-related, i.e, when the model training is completed, the parameters are pruned according to their values, which means that for different datasets, the pruned neural network is different; while some pruning methods perform pruning on the initialized model, such as the lottery ticket hypothesis~\cite{frankle2018the}, but it is still implemented by pruning after pre-training the original model, it uses the initialized parameters to reset the pruned model so that an initialized sub-network is found to represent most of the performance of the original model. In general, all these pruning methods are based on an iterative framework and all these methods need at least three steps to achieve pruning, in such process, specific dataset needs to be used for multiple rounds of training to finally obtain a pruned model. In this paper, we first raise a question: Can the pruning optimization be done from the perspective of the network structure itself? In other words, without using a dataset to specify model parameters, can we find a better sub-network of DNN model that can characterize most of its performance?

We answer this question from the perspective of graph theory with the information flow in the graph structure of neural networks. You et al.~\cite{you2020graph} proposed a method to map a DNN to a graph based on the division of neurons and the connection among these divisions, which is called the graph structure of neural network. They searched for different graph structures to find a neural network model with better performance under the same model complexity. In this paper, we conduct a further study on the graph structure of neural networks and use it for pruning neural networks. We first limit the search space of the graph structure to the range of regular graphs to significantly reduce the complexity of the search, and then propose a neural network pruning method based on the average shortest path length minimization of the regular graph.
In general, the main contributions of this paper are as follows:
\begin{itemize}
\item We implement the pruning of neural network from the perspective of graph structure and propose regular graph based pruning (RGP), which prunes neural network through its regular graph structure, and determines the pruning ratio of the neural network according to the node degree value of the graph.

\item We find that the average shortest path length of the graph is negatively correlated with the performance of the corresponding neural network. Therefore, a graph structure search algorithm based on the principle of minimizing average shortest path length is proposed, which can be used in the search space to obtain the graph with a better neural-network-performance.

\item Through the analysis of graphs and neural networks, we explain how the average shortest path length affects the performance of neural networks: The average shortest path length is negatively related to the back-propagation resistance of the output neuron, and reflects the number of parameters that can be affected by the gradient of each output neuron.

\item The experimental results on common models and common datasets show that the graph-guided neural network pruning framework can make DNN lose very little accuracy in the case of pruning at a very high ratio (more than 90\% on both parameters and FLOPs reduction). And compared to other neural network pruning methods based on iterative framework, our method is one-shot and the graph structure obtained by our method can be used in multiple models without re-searching.
\end{itemize}

\section{Related Work}
\label{research}
Existing works on pruning DNN can be roughly divided into three categories.

\textbf{Pruning before training} This kind of pruning is applied at initialization, and the mask keeps fixed during training. Frankle et al.~\cite{frankle2018the} articulated the lottery ticket hypothesis: a randomly-initialized, dense network contains a sub-network that when trained in isolation, can achieve comparable performance to the original network. Inspired by the lottery ticket hypothesis, many works have been done to prove it. Lee et al.~\cite{lee2018snip} introduced connection sensitivity to identify structurally important connections in the network for a specific task, then pruned the original network with the mask obtained by the connection sensitivity. Verdenius et al.~\cite{verdenius2020pruning} further proposed sensitivity criterion iteratively in smaller steps. Based on the lottery ticket hypothesis, Malach and Yehudai~\cite{malach2020proving} stated that for every bounded distribution, an over-parameterized DNN with random bounded weights contains a sub-network that can match the performance of the target network without any further training. Although the above methods apply pruning in the initialization process, the way of obtaining the mask still requires training with specific datasets. On the contrary, our approach belongs to pruning before training, but unlike previous works concentrating on regrading the importance of weights as the pruning criterion, we utilize graph theory as guidance to prune DNN and 
no training is required in this process. To the best of our knowledge, we are the first one who use graph theory to guide the pruning throughout the process.

\textbf{Pruning during training} This kind of pruning is a gradual pruning method proposed by Zhu et al.~\cite{zhu2017prune}, can be seamlessly incorporated within the training process. Zhang et al.~\cite{zhang2018systematic} formulated the pruning as a nonconvex optimization problem with combinatorial constraints meeting the requirements of the sparsity, then solved it with the alternating direction method of multipliers. He et al.~\cite{ijcai2018-309} came up with a soft filter pruning, which enabled pruned filters to be updated throughout training the model. Meng et al.~\cite{NEURIPS2020_ccb1d45f} introduced a brand new pruning paradigm named stripe-wise pruning, they utilized the filter skeleton (a learnable matrix) to efficiently learn the shape of each filter. Lin et al.~\cite{Lin2020Dynamic} proposed a dynamic pruning with feedback to reactivate prematurely pruned weights. Liu et al.~\cite{LIU2020Dynamic} presented dynamic sparse training with learnable layer-wise pruning thresholds, which could be adjusted dynamically via back-propagation.

\textbf{Pruning after training} Getting a sparse network usually needs three-step pipeline, including pre-training a dense network, pruning as well as fine-tuning~\cite{NIPS2015_ae0eb3ee}. Han et al.~\cite{han2016dsd} pruned the unimportant weights and then restored them to increase network capacity, finally fine-tuned the whole dense network. Lin et al.~\cite{ijcai2020-94} utilized the artificial bee colony algorithm in the pruning stage to efficiently find optimal pruned structure. Sehwag et al.~\cite{sehwag2020hydra} regarded pruning as an empirical risk minimization problem with adversarial loss objectives, then used three-stage pruning to simultaneously obtain remarkable benign and robust accuracy. Luo et al.~\cite{luo2020autopruner} combined pruning and fine-tuning into a single end-to-end trainable system.

Completely different from the above methods, we proposed a graph-based pruning, which utilizes the graph theory as guidance to find a sparsity mask to prune the network at initialization. Our method also demonstrates lottery ticket hypothesis. Compared to training a DNN with millions of parameters, finding a good sub-network (our method) is more effortless.

\begin{figure*} [!t]
\centering
\includegraphics[width=0.99\textwidth]{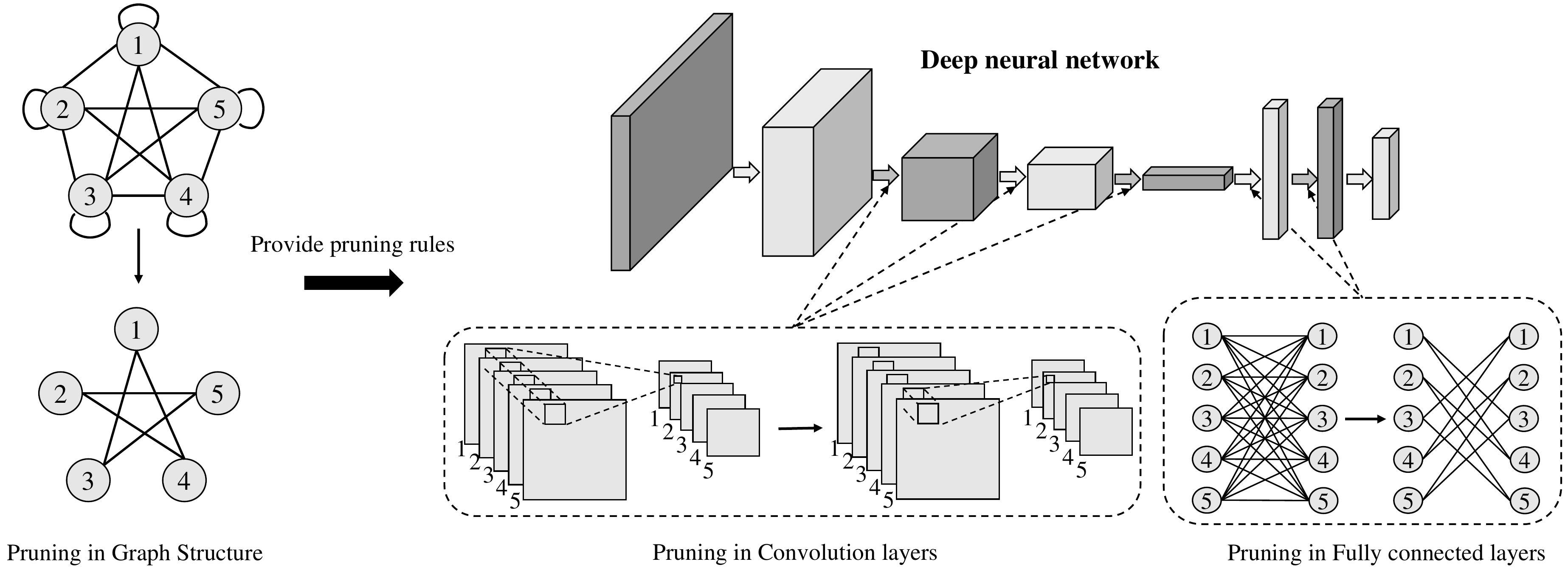}
\caption{The framework of graph-based pruning. Here each node of the graph represents part of each layer of neurons in the DNN.}
\label{Fig:framwork}
\end{figure*}

\section{Regular Graph Based Pruning of Neural Network}
\label{method}
In this section, we introduce the detail of regular graph based pruning of neural network.

\subsection{Graph pruning to DNN pruning}
The concept of graph structure of neural network was proposed by You et al.~\cite{you2020graph}, which can map a DNN containing any number of convolutional layers and any number of fully connected layers into a graph. The core is to divide the neurons in each fully connected layer and the convolution channels in each convolutional layer into $n$ equal parts. When the information is transmitted from
one layer to the next layer, the $n$ equal parts of the previous layer will be connected to the $n$ equal parts of the next layer, leading to connected edges in an $n$-node undirected homogeneous graph. 
Similarly, an undirected homogeneous graph can also be mapped into a neural network according to the same rules. Formulaically, let $f_l(\cdot)$ be the fully connected operation, and $f_c(\cdot)$ be the convolution operation, a traditional fully connected layer operation can be expressed as:
\begin{equation}
X_{i+1}=f_l(X_{i}),  \label{EQ:1}
\end{equation}
where $X_i=\{x_i^1, x_i^2, ..., x_i^p\}$ and $X_{i+1}=\{x_{i+1}^1, x_{i+1}^2, ..., x_{i+1}^q\}$ represent the set of neurons in the $i$th layer and that in the $(i+1)$th layer, respectively. Then, neurons of each layer is divided into $n$ equal parts: $X_i=\{X_i^1, X_i^2, ..., X_i^n\}$ and $X_{i+1}=\{X_{i+1}^1, X_{i+1}^2, ..., X_{i+1}^n\}$ where $X_i^1=\{x_i^1,...,x_i^{p/n}\}$ (if $n$ is not divisible by the number of neurons, the number of neurons in the last part will be less than the other parts, i.e, $len(X_i^1)=...=len(X_i^{n-1})>len(X_i^n)$). Then we have an undirected graph with $n$ nodes:  $G=(V, E)$, where $V=\{v_1, v_2, v_3, ..., v_n\}$ and $E$ is the set of edges. Then, according to the edge distribution in $E$, we can get the pruned fully connected layer: 
\begin{equation}
X_{i+1}=\{X_{i+1}^1, X_{i+1}^2, ..., X_{i+1}^n\}, X_{i+1}^j=\sum\limits_{k}^{v_k \in N(v_j)} {f_l(X_i^k)},  \label{EQ:2}
\end{equation}
where $N(v_j)$ is the collection of $v_j$'s neighbors in the graph. For the convolutional layer, similar operations can be performed in the channel dimension, and we can get:
\begin{equation}
C_{i+1}^j=\sum\limits_{k}^{v_k \in N(v_j)} {f_c(C_i^k)},  \label{EQ:3}
\end{equation}
where $C_i^k$ and $C_{i+1}^j$ respectively represent the $k$th part of feature map channels of the $i$th convolutional layer and the $j$th part of feature map channels of the $(i+1)$th convolutional layer.

Based on the graph structure of neural network, we proposed a pruning framework of DNN, as shown in  Figure~\ref{Fig:framwork}. Firstly, we choose a DNN as the original network to be pruned. Secondly, an appropriate complete graph is used to represent the original neural network, here the “appropriate” means the number of the nodes of the graph is less than the number of neurons or convolution channels of each layer (except input layer and output layer), so that each node of the graph can represent part of the DNN layer. Thirdly, a large number of edges in the graph will be removed according to the pruning ratio. Finally, in the DNN structure, the existence of parameters of each convolutional layer and fully connected layer will be decided according to the edges in the graph to finish pruning. In actual operation, we only need to generate a sparse graph structure to complete the pruning of the DNN, instead of starting optimization from the complete graph.

Since the graph structure directly represents the existence of the parameters of each layer in neural network, the selection and search optimization of the graph structure are critical to the performance of the pruned network. Next, we will introduce the strategy for selecting graph structure.

\subsection{Regular graph as the search space}
After we determine the number of nodes of a graph, the potential search space is huge, since there are so many kinds of graphs. Here, we choose regular graph as the target graph to greatly reduce the search space, the reasons are as follow: 

Firstly, a regular graph is a graph where each node has the same number of neighbors (has the same degree), which is reflected in the neural network structure that each neuron of a fully connected layer gets value from the same number of neurons in the previous layer, then, is connected to the same number of neurons in the next layer; and so as each channel of a convolutional layer's feature map in the convolution operation. If we rewrite Eq.(\ref{EQ:2}) and Eq.(\ref{EQ:3}) into a form with parameters as:
\begin{equation}
X_{i+1}^j=\sum\limits_{k}^{v_k \in N(v_j)} {f_l(X_i^k, W_i^k)}; C_{i+1}^j=\sum\limits_{k}^{v_k \in N(v_j)} {f_c(C_i^k, F_i^k)}, \label{EQ:4}
\end{equation}
where $W_i^k$ and $F_i^k$ represent the fully connected parameters and convolution kernel parameters, respectively, then a regular graph means for all $j$, $X_{i+1}^j$ is obtained by the operation with the same number of $W_i^k$, and $C_{i+1}^j$ is obtained by the operation with the same number of $F_i^k$. Such structure shows the equal importance of each neuron or convolution channel in their own layers from the perspective of the model topology, and we think this is appropriate because before the model is initialized and trained, the neurons or convolution channels in each layer may need to be equally important so that each neuron or channel has the same probability of being trained as an important neuron or channel in subsequent training. This principle is also in line with the topology of traditional dense neural networks, i.e, the original model structure can be mapped as a special regular graph (a complete graph with self-loops). 

Secondly, in network science, regular graph is an important type of network. Its strictly controlled degree distribution allows researchers to deeply study other properties of the network. The space of regular graph contains some networks with extremely good properties, such as entangled network~\cite{Donetti2005EntangledNS}. Actually, the graph structure we searched is very close to the entangled network, one can see more details in Appendix.

Finally, for pruning implementation, a regular sparse structure (each neuron uses the same number of weights in a layer) can easily achieve speedup through dense computation, while other graphs based sparse structure may do it a bit difficult. (A more detailed discussion can be seen in Appendix.)

We also compare the performance of the neural network corresponding to the regular graph and the random graph under the same sparsity through experiments (see Appendix). The experimental results also show that under the same sparsity, the performance of the pruning model based on random graphs is not as good as that based on regular graphs.

\subsection{Optimization of regular graph via minimizing average shortest path length}
In a graph, average shortest path length (ASPL) is defined as the average number of steps along the shortest paths for all possible pairs of nodes. ASPL facilitates the quick transfer of information and reduce costs in a real network such as Internet.

Here, with a small ASPL graph, a neural network structure with more efficient gradient back-propagation can be obtained. We iteratively minimize ASPL of a regular graph. As shown in Algorithm~\ref{alg:1}, firstly, we randomly select two edges in the initial graph to exchange, and then determine whether the ASPL of the entire graph is reduced. If it is reduced, keep this change, otherwise select two other edges again and repeat the above exchange process, we set a large enough number $m$ as the number of repetitions of the above process, in our actual experiments, $m=10000$.

\begin{algorithm}[!h]
    \renewcommand{\algorithmicrequire}{\textbf{Input:}}
    \renewcommand{\algorithmicensure}{\textbf{Output:}}
    \caption{Minimize ASPL of Regular Graph}
    \label{alg:1}
    \begin{algorithmic}[1]
        \REQUIRE
        A regular graph: $G=(V, E)$, where $V=\{v_1, v_2, v_3, ..., v_n\}, E=\{(v_i, v_j), ...,\}, (i,j \in (1,2, ..., n))\}$; Positive integer: $m$
        \ENSURE
        A regular graph: $G'$.
        \WHILE {$m > 0$}
        \STATE $G=(V, E)$;
        \STATE Randomly select  edge pair in $E$: $\{(v_i, v_j), (v_p, v_q)\}$, where $(i,j,p,q \in (1,2, ..., n))$; \\
        \STATE Swap as $E'=E-\{(v_i, v_j), (v_p, v_q)\}+\{(v_i, v_p), (v_j, v_q)\}$; \\
        \STATE $G'=(V, E')$;
        \IF {$ASPL(G') \leq ASPL(G)$} 
            \STATE $E=E'$; \\
        \ELSE  
            \STATE $E=E; G'=G$; \\
        \ENDIF
        \STATE $m=m-1$; \\
        \ENDWHILE 
        \RETURN $G'$.
    \end{algorithmic}
\end{algorithm}

Experimental results show that when the degree value of the graph and the number of nodes remain unchanged, the smaller the ASPL, the better performance neural network will have.

\begin{figure*} [!t]
\centering
\includegraphics[width=0.99\textwidth]{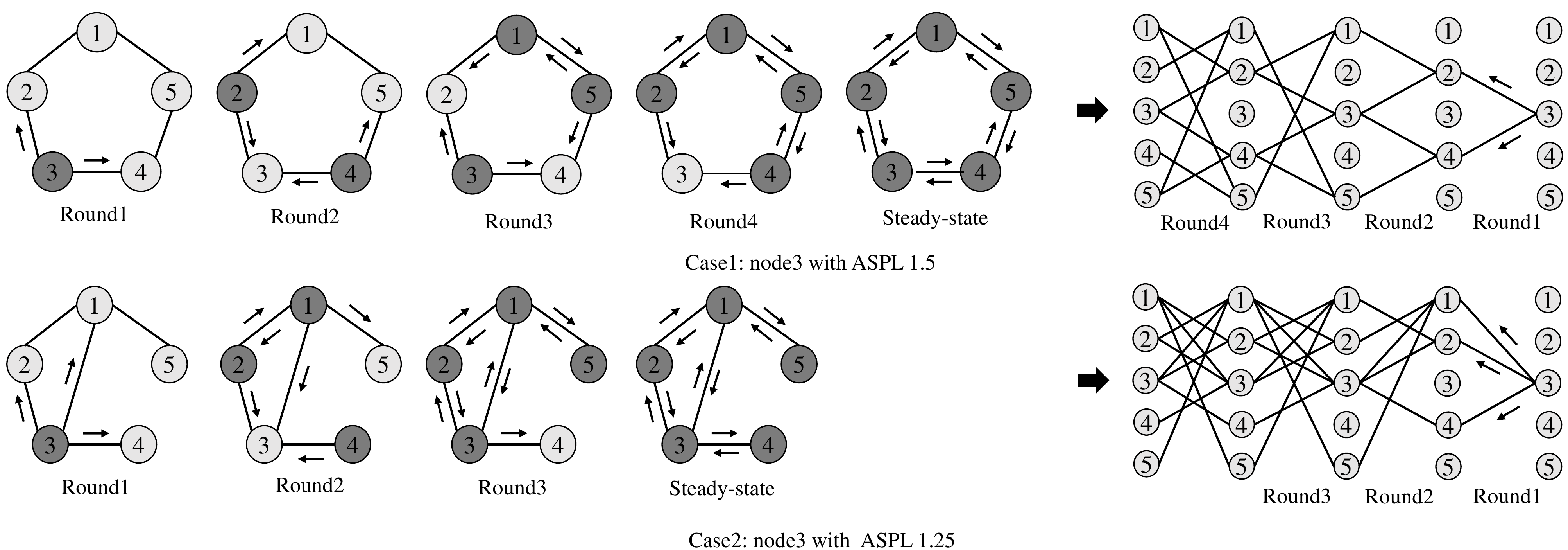}
\caption{A simple example to illustrate information transfer in graph and gradient transfer in neural network. We analyze node 3 and set two cases of different ASPL.}
\label{Fig:analysis}
\end{figure*}

\begin{figure*} [!t]
\centering
\includegraphics[width=0.99\textwidth]{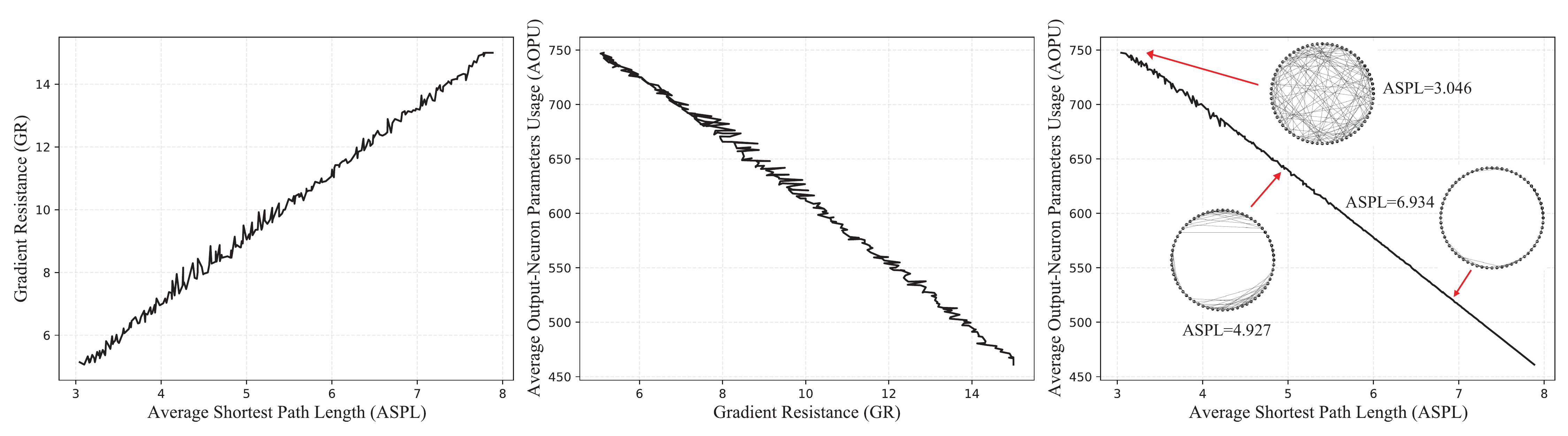}
\caption{The statistical relationship between ASPL, GR and AOPU. We calculated three values on 64-node, 4-degree regular graphs, the corresponding neural network is a 15-layer MLP, and each layer contains 64 neurons.}
\label{Fig:analysis2}
\end{figure*}

\subsection{How can the ASPL affect the performance of neural networks}
\label{how_affect}

\begin{figure*} [!t]
\centering
\includegraphics[width=0.99\textwidth]{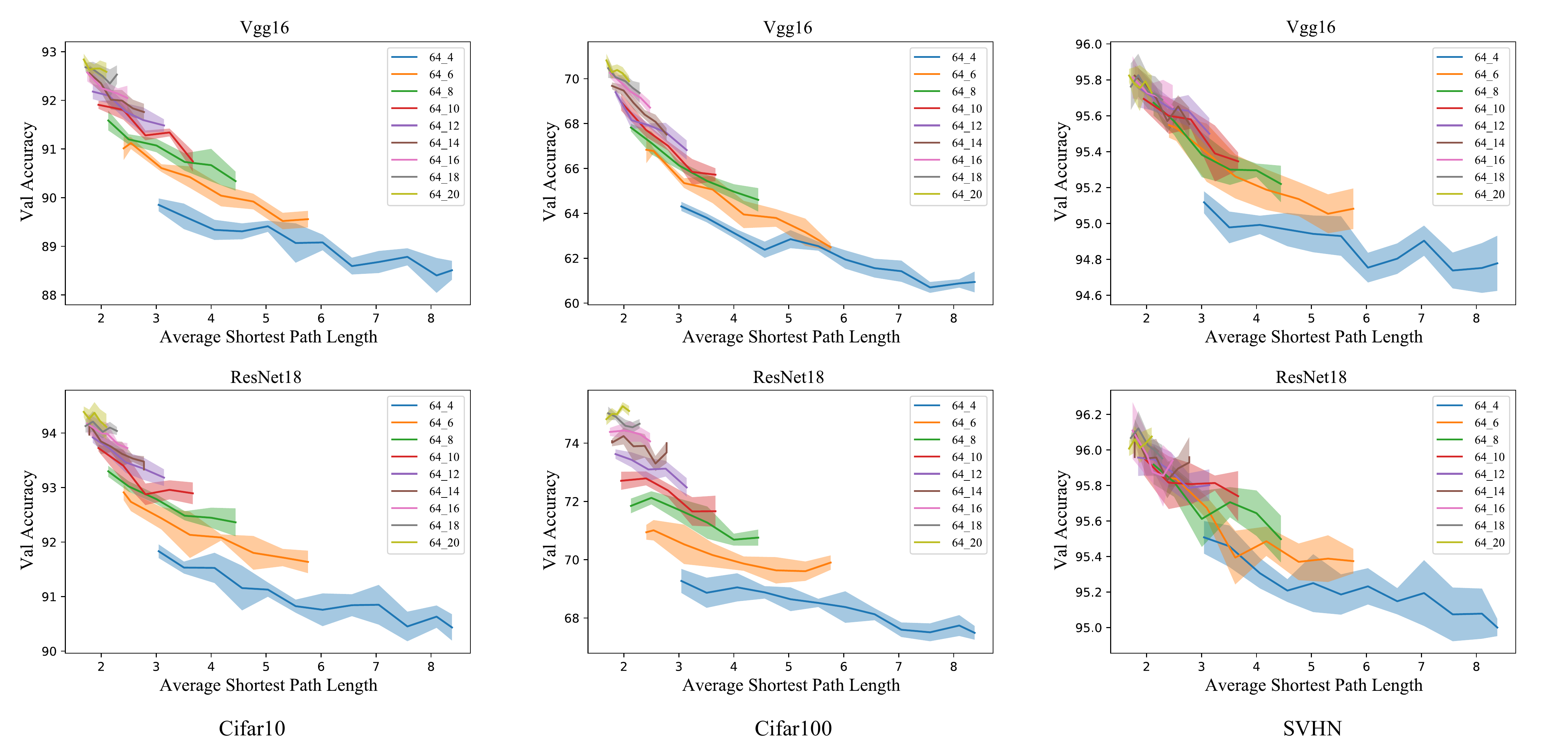}
\caption{Performance changes during graph structure search. We set different node degree values to represent the pruning ratio of the model. For example, $64\_4$ represents the degree value of each node is $4$, and the corresponding pruning ratio is about 93.75\%.}
\label{Fig:result1}
\end{figure*}

Here we analyze how the ASPL in the graph structure affects the performance of the neural network. We use a graph with 5 nodes as an example, as shown in Figure~\ref{Fig:analysis}. For simplicity, the neural network corresponding to the graph structure is set up as a 5-layer MLP, and each layer contains 5 neurons. We analyze node 3 and set up two cases: In case 1, the ASPL of node 3 is 1.5, in case 2, the ASPL of node 3 is 1.25. The calculation of each layer is represented as each round of information transmission in the graph structure. From the perspective of graph, we set that each node will send messages to its neighbors in the next round after receiving neighbors' messages, at the initial moment, node 3 generates a message and passes it to its neighbors, and then this message will continue to spread in the graph. Finally, in a certain round, all nodes will receive new information from their neighbors and then will keep sending and receiving status, we call such round steady-state, as shown in Figure~\ref{Fig:analysis} left. We can see in case 1, it needs 4 rounds to get the steady-state while in case 2 it only needs 3 rounds. So how does this phenomenon behave in neural networks? We look at the neural network from the perspective of back-propagation, as shown in Figure~\ref{Fig:analysis} right. For case 1, the gradient information of neuron 3 in the output layer will pass through 4 layers. In the fourth layer (the first layer from left), it affects all neurons in this layer. For case 2, the gradient information of neuron 3 in the output layer only needs to pass through 3 layers to affect all neurons in a certain layer. This corresponds to the number of rounds that make graph reach the steady-state. We call such number Gradient-Resistance (GR) of output neuron 3 or node 3, it measures the difficulty of the gradient starting from neuron 3 in the last layer to reach all neurons in a certain layer. Further, we can find that a smaller GR of node 3 means there will be more entire layers that can be used to get the result of neuron 3 in the output layer. It is more conducive to making the output neuron 3 get more powerful prediction or classification ability, because more parameters are used to get the value of output neuron 3. For all nodes of a graph, we can calculate its average GR as the GR of the graph.

The above is an analysis on a simple graph. In fact, for the regular graphs used in our method, we calculate the GR value of each graph, as shown in Figure~\ref{Fig:analysis2}, we choose 64-node regular graphs with each node of degree 4, the ASPL ranges from about 3 to 8, and we set the corresponding neural network as a 15-layer MLP, with 64 neurons in each layer. We also calculate the average output-neuron parameters usage (AOPU), which means the average of the amount of parameters used to calculate the value of each output neuron. We can see that ASPL and GR have approximately linear positive correlation, ASPL and AOPU have approximately linear negative correlation, which statistically validates our above analysis conclusion. In summary, a small ASPL graph means that at the same sparsity, more parameters are used to calculate the final neuron output. Or we can say, under the condition that the total amount of model parameters remains unchanged, a small ASPL graph improve the reuse rate of model parameters.

\section{Experiments and Results}
\begin{table*}
  \caption{Pruning results of RGP on three datasets with different pruning ratio.}
  \label{table1}
  \centering
  \resizebox{0.7\linewidth}{!}{\large
  \begin{tabular}{ccccccc}
    \toprule
    Dataset & Model & Top1 Acc & Acc Drop & Parameters & \makecell[c]{Parameters \\Reduction} & \makecell[c]{FLOPs \\Reduction} \\
    \midrule
    \multirow{10}*{Cifar10} & \multirow{5}*{VGG16} & 93.14\% & 0.00\% & 15.24M & 0.00\% & 0.00\% \\
    & & 92.91\% & 0.23\% & 4.76M & 68.77\%(64\_20) & 68.65\%(64\_20) \\
    & & 92.76\% & 0.38\% & 3.81M & 75.00\%(64\_16) & 74.89\%(64\_16) \\
    & & 92.17\% & 0.97\% & 2.38M & 84.38\%(64\_10) & 84.25\%(64\_10) \\
    & & 91.45\% & 1.69\% & 1.43M & 90.62\%(64\_6) & 90.49\%(64\_6) \\
    \cmidrule(r){2-7}
    & \multirow{5}*{ResNet18} & 94.73\% & 0.00\% & 11.16M & 0.00\% & 0.00\% \\
    &  & 94.53\% & 0.20\% & 3.49M & 68.73\%(64\_20)  & 68.67\%(64\_20)  \\
    &  & 94.38\% & 0.35\% & 2.79M & 75.00\%(64\_16) & 74.92\%(64\_16) \\
    & & 93.70\% & 1.03\% & 1.74M & 84.41\%(64\_10) & 84.28\%(64\_10) \\
    & & 93.05\% & 1.68\% & 1.05M & 90.59\%(64\_6) & 90.52\%(64\_6) \\
    \midrule
    \multirow{10}*{Cifar100} & \multirow{5}*{VGG16} & 71.83\% & 0.00\% & 15.29M & 0.00\% & 0.00\% \\
    & & 71.15\% & 0.68\% & 4.78M & 68.74\%(64\_20) & 68.65\%(64\_20) \\
    & & 70.40\% & 1.43\% & 3.82M & 75.02\%(64\_16) & 74.89\%(64\_16) \\
    & & 69.33\% & 2.50\% & 2.39M & 84.37\%(64\_10) & 84.25\%(64\_10) \\
    & & 67.42\% & 4.41\% & 1.43M & 90.65\%(64\_6) & 90.49\%(64\_6) \\
    \cmidrule(r){2-7}
    & \multirow{5}*{ResNet18} & 77.04\% & 0.00\% & 11.21M & 0.00\% & 0.00\% \\
    &  & 75.28\% & 1.76\% & 3.50M & 68.78\%(64\_20) & 68.67\%(64\_20) \\
    &  & 74.80\% & 2.24\% & 2.80M & 75.02\%(64\_16) & 74.92\%(64\_16) \\
    & & 72.96\% & 4.08\% & 1.75M & 84.39\%(64\_10) & 84.28\%(64\_10) \\
    & & 71.51\% & 5.53\% & 1.05M & 90.63\%(64\_6) & 90.52\%(64\_6) \\
    \midrule
    \multirow{10}*{SVHN} & \multirow{5}*{VGG16} & 96.09\% & 0.00\% & 15.24M & 0.00\% & 0.00\% \\
    & & 95.91\% & 0.18\% & 4.76M & 68.77\%(64\_20) & 68.65\%(64\_20) \\
    & & 95.97\% & 0.12\% & 3.81M & 75.00\%(64\_16) & 74.89\%(64\_16) \\
    & & 95.77\% & 0.32\% & 2.38M & 84.38\%(64\_10) & 84.25\%(64\_10) \\
    & & 95.63\% & 0.46\% & 1.43M & 90.62\%(64\_6) & 90.49\%(64\_6) \\
    \cmidrule(r){2-7}
    & \multirow{5}*{ResNet18} & 96.03\% & 0.00\% & 11.16M & 0.00\% & 0.00\% \\
    &  & 96.14\% & -0.11\% & 3.49M & 68.73\%(64\_20) & 68.67\%(64\_20) \\
    &  & 96.22\% & -0.19\% & 2.79M & 75.00\%(64\_16)  & 74.92\%(64\_16)  \\
    & & 96.03\% & 0.00\% & 1.74M & 84.41\%(64\_10) & 84.28\%(64\_10) \\
    & & 95.86\% & 0.17\% & 1.05M & 90.59\%(64\_6) & 90.52\%(64\_6) \\
    
    \bottomrule
  \end{tabular} }
\end{table*}
In this section, we do lots of experiments to verify the effectiveness of our method.

\subsection{DNN models and datasets}
\label{implement}

In this paper, we implemented our method on multiple common DNN models and public datasets. The DNN models used include ResNet18/56~\cite{he2016deep}, VGG16~\cite{simonyan2014very}. The datasets used include Cifar10, Cifar100~\cite{krizhevsky2009learning} and SVHN~\cite{netzer2011reading}.  We set the regular graphs of different node degrees according to the pruning ratio from high to low, and search from the nearest neighbor graph to the target graph step by step according to the ASPL minimum principle. Then the target graph is mapped to the sub-network of different DNN models. And we use multiple datasets for training and performance testing. 

In our experiments, for ResNet18 and VGG16, the models are represented as 64-nodes regular graphs; for ResNet56, due to the limitation of the number of convolution channels, the model is represented as 16-nodes regular graphs. For evaluation, we use number of parameters and FLOPs (Float Points Operations) to evaluate the model size and computational requirement. For configurations, we use Pytorch~\cite{paszke2017automatic} to implement all models. Stochastic Gradient Descent algorithm (SGD) with an initial learning rate of 0.1 and weight decay of 0.0005 is used as the optimization strategy. And the batch size is set to 256. We set a total training epoch of 100 to ensure that the models are fully trained, and for each model we choose the highest test accuracy as the classification performance of the model. Training and testing of all models are performed on the GPUs of NVIDIA Tesla V100.

\subsection{Performance changes during graph structure search}
We first verify the effect of the change of ASPL on the model performance in the process of graph search optimization. We start with the nearest neighbor graph and minimize the ASPL of the graph by randomly exchanging edge pairs which is shown in Algorithm~\ref{alg:1}. More specifically, we set different node degree values to represent the pruning ratio of the model, the degree ranges from 4 to 20. For example, if we want to keep about $4/64$ (6.25\%) of neural network parameters through pruning, then we start with a 4-nearest neighbor graph with 64 nodes, and minimize the ASPL of the graph through Algorithm~\ref{alg:1}, 
finally, the obtained graph is mapped to a specified neural network to complete the pruning. We use three public datasets Cifar10, Cifar100 and SVHN to train and test the pruned model, and repeat the experiment for many times to reduce the chance, the target original models are VGG16 and ResNet18.

The experimental results are shown in Figure~\ref{Fig:result1}, where $64\_4$ and $64\_20$ represents that degree value of each node is $4$ and $20$ in the graph, respectively. As we can see, for all the results, when more parameters of a model are pruned, the validation accuracy of the model decreases steadily, from $64\_20$ to $64\_4$. Under the same pruning ratio, the classification performance of the model and the ASPL of the graph structure show an obvious negative correlation. 

\begin{table*}
  \caption{Comparison results with other methods on Cifar10.}
  \label{table2}
  \centering
  \resizebox{0.7\linewidth}{!}{\large
  \begin{tabular}{cccccc}
    \toprule
     Dataset & Model & Method & Top1 ACC & Parameters Reduction & FLOPs Reduction \\
    \midrule
    \multirow{11}*{Cifar10} 
    & \multirow{6}*{VGG16} & GAL-0.05~\cite{lin2019towards}  & 92.03\% & 77.6\% & 39.6\% \\
    & & HRank1~\cite{lin2020hrank} & 92.34\% & 82.1\% & 65.3\% \\
    & & \textbf{RGP1(ours)} & 92.39\% & 81.2\%(64\_12) & 81.1\%(64\_12) \\
    & & GAL-0.1~\cite{lin2019towards} & 90.73\% & 82.2\% & 45.2\% \\
    & & HRank2~\cite{lin2020hrank} & 91.23\% & 92.0\% & 76.5\% \\
    & & TRP~\cite{xu2020trp} & 91.62\% & - & 77.82\% \\
    & & \textbf{RGP2(ours)} & 91.45\% & 90.5\%(64\_20) & 90.6\%(64\_20) \\
    \cmidrule(r){2-6}
    & \multirow{4}*{ResNet56} & He et al.~\cite{he2017channel} & 90.80\% & -  & 50.6\% \\
    &  & GAL-0.8~\cite{lin2019towards} & 90.36\% & 65.9\% & 60.2\% \\
    &  & HRank~\cite{lin2020hrank} & 90.72\% & 68.1\%  & 74.1\% \\
    & & \textbf{RGP(ours)} & 91.50\% & 75.3\%(16\_4)  & 74.5\%(16\_4) \\
    \bottomrule
  \end{tabular} }
\end{table*}

\subsection{Performance of pruned models}
Under the same pruning ratio, we choose the model with the highest classification accuracy as the final performance of the pruned model, and record the sub-network (pruned model) performance under 4 different pruning ratios, as shown in Table~\ref{table1}. As we can see, our method can greatly reduce the complexity of the model in terms of parameters and FLOPs at the same time, and the ratio of parameter reduction is almost equal to the ratio of FLOPs reduction. This is due to the fact that our graph structure guided pruning method has the same pruning rules for each layer of the neural network. As far as we know, most of the existing pruning strategies can often reduce a large amount of parameters (more than 80\%), but relatively, the reduction ratio of FLOPs is far less than the reduction ratio of the parameter amount (often less than 70\%). Our method can reduce the number of parameters and FLOPs by more than 90\% at the same time, but only reduce the top1 accuracy of a very small amount (1.69\% of VGG16 on Cifar10, 0.46\% of VGG16 on SVHN and 0.17\% of ResNet18 on SVHN).

\subsection{Comparison with other neural network structure pruning methods}
We also compare our results with other neural network pruning methods. However, most results of other methods are not with a high pruning ratio (less than 70\%) and the advantage of our method is mainly to maintain good model performance under high pruning ratio, therefore, we choose several neural network pruning methods under the high pruning ratio for comparison, as shown in Table~\ref{table2}. We compare these methods on the dataset Cifar10, the method we are more concerned about is HRank~\cite{lin2020hrank}, because as an advanced pruning method, it has detailed performance records and complete comparison results with other methods. We list the results of the maximum and the second-maximum pruning ratios in HRank, noted as HRank1 and Hrank2, respectively. Correspondingly, we choose the results of two different pruning ratios of RGP (as RGP1 and RGP2) to compare with the two results of HRank. Also, several other methods are listed, too. As we can see, for VGG16, our method (RGP2) can achieve an accuracy of 91.45\% when both the parameter amount and FLOPs are reduced by more than 90\%, while HRank2 achieves an accuracy of 91.23\% when the parameter amount is reduced by 92\% but FLOPs is only reduced by 76.5\%. RGP1 is also better than HRank1 in pruning ratio and accuracy. TRP~\cite{xu2020trp} is slightly higher than RGP2 in accuracy but its FLOPs reduction is still significantly smaller than ours. For ResNet56, our method shows a greater pruning ratio and better top1 accuracy than other methods. In general, the advantage of our method is that it can form a DNN sub-network at one time through the mapping of the graph structure, and its parameters and FLOPs can be greatly reduced at the same time, and better performance can be obtained at a high pruning ratio than other methods.

\section{Conclusion and Broader Impact}
\label{Impact}
In this paper, we propose RGP, a parameter pruning framework based on the regular graph structure of the neural network. The framework can change the degree value of the regular graph to set the pruning ratio, and find a sub-network with better performance under a certain pruning ratio by minimizing ASPL. We also explain the negative correlation between ASPL and neural network performance through analysis and statistical calculations of output neurons. The graph-guided pruning becomes an efficient method that can be directly applied to multiple models without re-searching the graph structure and shows a strong precision retention capability with extremely high parameter reduction and FLOPs reduction. In the future, we may explore the impact of more graph topologies on model performance since there are so many graph categories, and a more suitable neural network to graph mapping may be explored to form an intersecting study of the two fields.

Our work can be used to achieve large-scale compression of neural networks, and can inspire researchers in the graph field and network pruning field to implement model pruning from the perspective of graph structure. They can look for sub-networks of neural networks from the topology rather than through iterative pruning, and such sub-networks may be more explanatory. The disadvantage of our method is that when the pruning ratio is not high, it may lose more accuracy compared to other iterative pruning methods. Also, the regular graphs may not be the best choice, more graph structures may need to be explored and analyzed to establish a better relationship between neural networks and graphs.

{\small
\bibliographystyle{ieee_fullname}
\bibliography{RGP}
}


\clearpage

\appendix

\section{Appendix}

\subsection{The existence of GR value}
We discuss the existence of GR value in Section~\ref{how_affect}, because for some graph structures, information from a certain node may not be able to spread throughout the entire graph in a certain round, which means the GR value is infinite.

First, we make a supplement to Algorithm~\ref{alg:1}. In actual operation, we have tested the connectivity of the graph after each time of edges' swap, and the disconnected graph will be deleted to ensure that the searched graph is connected. We prove that under the condition of connected graph, when there is always an even-length path (can be repeated) from a node to other nodes (including the node itself) or there is always an odd-length path from a node to other nodes, the node has a finite GR value. The proof is as follows:

Suppose that $a$ is a node of a connected graph $G$, $\{a+n\}$ represents the set of all nodes in $G$ whose path length (may not be the shortest path length) to $a$ is $n$, for example $\{a+1\}$ represents the set of all neighbors of $a$. Assuming that the initial information is sent from $a$, the nodes that receive the information in each round can be expressed as follows:
\begin{equation}
\left\{
             \begin{array}{lr}
             \{a+1\}, &  Round=1, \\
             a \cup \{a+2\}, & Round=2, \\
             \{a+1\} \cup \{a+3\}, & Round=3, \\
             a \cup \{a+2\} \cup \{a+4\}, & Round=4. \\
             ...
             \end{array}
\right.
\label{EQ:5}
\end{equation}
More generally, Eq.(\ref{EQ:5}) can be written as:
\begin{eqnarray}
\left\{
             \begin{array}{lr}
             \{a+1\} \cup \{a+3\} \cup \dots \cup \{a+Round\}, (Round \\ \leq d\ \& \ Round\ is\ odd); \\
             a \cup \{a+2\} \cup \{a+4\} \cup \dots \cup \{a+Round\}, \\ (Round  \leq d\ \& \ Round\ is\ even) \\
             \end{array}
\right.
\label{EQ:6}
\end{eqnarray}
where $d$ represents the maximum value of the shortest path length between $a$ and other nodes. Then, we can see that when $Round\geq d$, in the odd round, all nodes with an odd path length to $a$ will receive the information at the same time. In the even round, all nodes with an even path length to $a$ will receive the information at the same time.

In our experiment, we did not encounter a situation where the GR value does not exist. This may be due to the setting of the regular graph so that all nodes can have several neighbors, which makes each node have a very high probability that there are paths with odd lengths and paths with even lengths to other nodes at the same time, especially when the graph is large.

\subsection{The convergence of Algorithm 1}
\label{Algorithm 1}
\begin{figure*} [!t]
\centering
\includegraphics[width=0.99\textwidth]{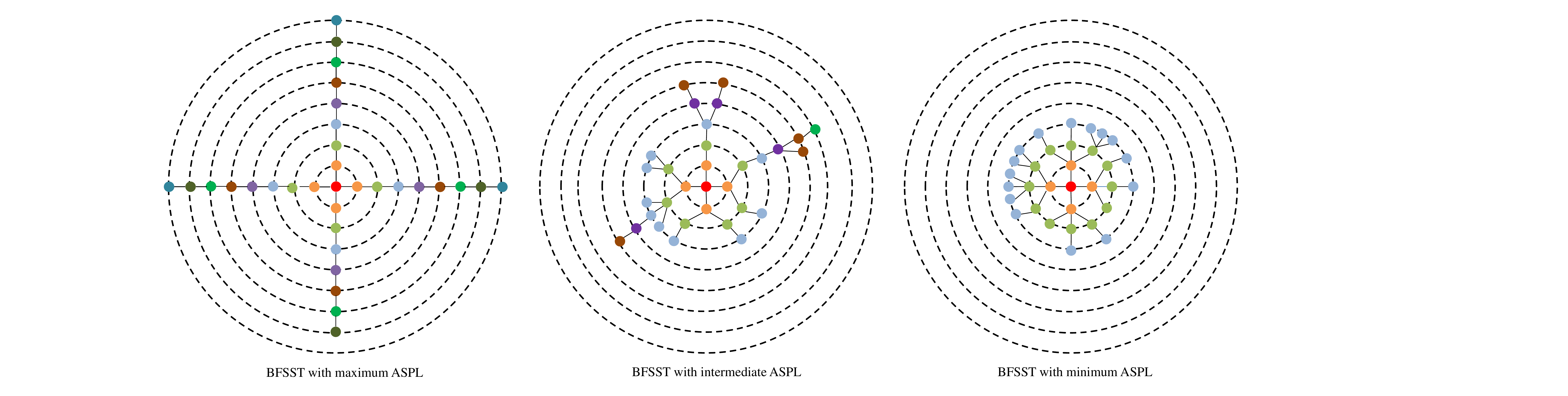}
\caption{The broad-first search spanning trees (BFSSTs) with different ASPL for different 32-nodes 4-degree regular graphs.}
\label{Fig:bfsst}
\end{figure*}

We discuss the search results of Algorithm~\ref{alg:1}. It needs to be mentioned that in the search process of Algorithm 1, we maintain the connectivity of the graph, and disconnected graphs will not be used as search results and intermediate results. 
For regular graphs, we use their broad-first search spanning trees (BFSSTs) to calculate the theoretical lower bound of their ASPL. For a regular graph to its BFSST, we first randomly choose a node as the root node of the tree, then starting from the root node, the neighbors of each node that are not searched will become the next layer of leaves of the node until the entire graph is searched. The ASPL of a graph can be calculated as:
\begin{equation}
L=\frac{\sum_{i=1}^NL_i}{N}  \label{EQ:A1}
\end{equation}
where $L_i$ is the ASPL between the root node $i$ and the other part in corresponding BFSST. And for 32\_4 regular graph, BFSST with maximum, intermediate, minimum ASPL is shown in Figure~\ref{Fig:bfsst}, respectively.

Then for regular graph with $N$ nodes and $k$ uniform degree, when each BFSST has the minimum ASPL, the graph has the minimum ASPL, thus we can calculate the theoretical lower bound of ASPL of the regular graph as:
\begin{equation}
\begin{aligned}
& L_{lower} = min(L_i) = (\sum_{i=1}^\theta ki(k-1)^{(i-1)}+ \\ & (\Theta +1)(N-1-\sum_{i=1}^\theta {k(k-1)^{(i-1)}})/(N-1) \\
& = \frac{\sum_{i=1}^\theta {ki(k-1)^{(i-1)}} + (\theta +1)(N+\frac{2-k(k-1)^\theta}{k-2})}{N-1} 
 \end{aligned}
  \label{EQ:A2}
\end{equation}
where $\theta$ represents the number of layers full filled by nodes in BFSST (for example, in figure~\ref{Fig:bfsst} right, it has two layers (circles) around the root node full filled with other nodes ), and is calculated as: 
\begin{equation}
\theta = \frac{ln((N-1)(k-2)/k+1)}{ln(k-1)}
\end{equation}

\begin{figure} [!t]
\includegraphics[width=0.49\textwidth]{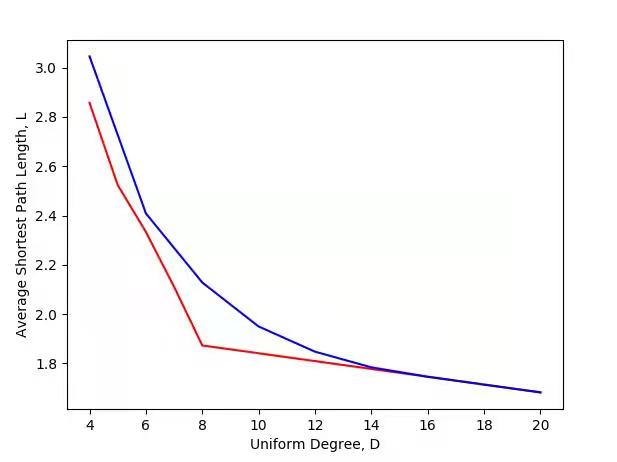}
\caption{Comparison of ASPL obtained by Algorithm~\ref{alg:1} with the theoretical lower bound, red for lower bound and blue for Algorithm~\ref{alg:1} ($m=10000$), for 64-nodes regular graphs.(Here uniform degree D is k in text.)}
\label{Fig:lb}
\end{figure}
We calculated the $L_{lower}$ of the regular graph with 64-nodes of different degrees, and compared it with the minimum ASPL obtained by our algorithm after 10,000 searches, as shown in Figure~\ref{Fig:lb}. We can see that when the degree value gradually increases, the theoretical lower bound of ASPL gradually decrease but the speed of change is getting slower and slower. When the degree value changes from 7 to 8, $\theta$ changes from 2 to 1, and then remains at 1, the change of lower bound of ASPL becomes very slow and the lower bound gradually overlaps with our search algorithm as the degree increase. Therefore, in most cases, the ASPL obtained by our search algorithm is close to the theoretical lower bound. When the degree value is large, our algorithm can reach the lower bound of ASPL.

In fact, we are excited to find that such optimal graph is very close to the entangled network~\cite{Donetti2005EntangledNS} in physics (or cage graph in mathematics) which shows good symmetry in terms of short average distances, large loops, and poor modularity, and exhibits an excellent performance such as robustness against errors and attacks, minimal first-passage time of random walks, good searchability, efficient communication, etc.

\subsection{The dense computation for regular sparse structure speedup}
\begin{figure*} [!t]
\centering
\includegraphics[width=0.99\textwidth]{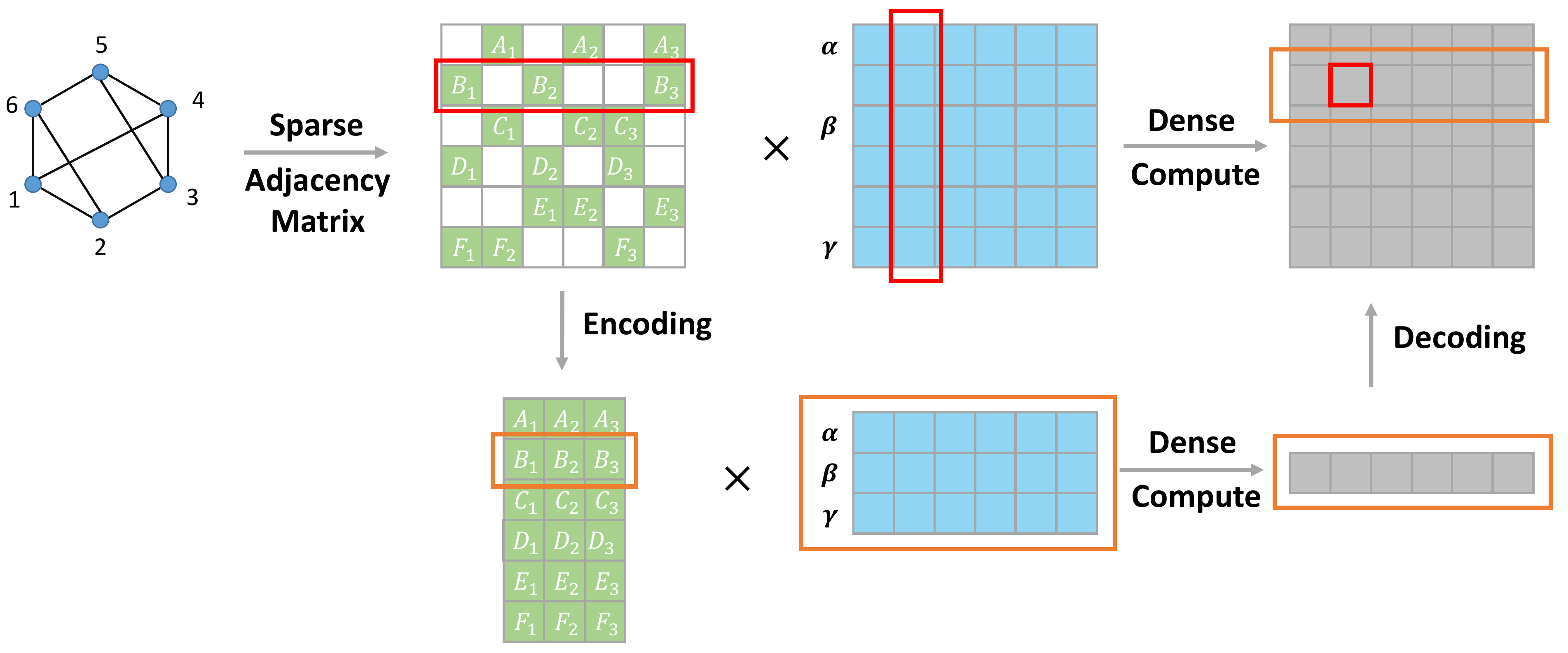}
\caption{A regular graph based NN can achieve a real speedup through dense computation, however, other graphs based NN cannot use the same dense computation framework because of the different degree values of each node.}
\label{Fig:speedup}
\end{figure*}

Actually, at present, the methods of weight pruning are mainly realized by multiplying the sparse mask with the weight matrix, and such methods are usually unable to obtain real acceleration in the deep learning framework based on Pytorch~\cite{paszke2017automatic} or Tensorflow~\cite{tensorflow2015-whitepaper}, because in these frameworks, floating-point multiplication with 0 cannot be omitted.
In the research of Lin et al.~\cite{lin20211}, they pointed out that regular parameters can be truly accelerated by encoding. For the regular graph based pruning, because each node has the same degree value, it is easier to achieve real acceleration through dense computation. One theoretical framework can be seen in Figure~\ref{Fig:speedup}, for the parameters of each layer, we can densify the sparse parameters through encoding, then implement the operation through the multiplication of the dense matrices, and finally restore the output through decoding.

\begin{figure} [!t]
\includegraphics[width=0.49\textwidth]{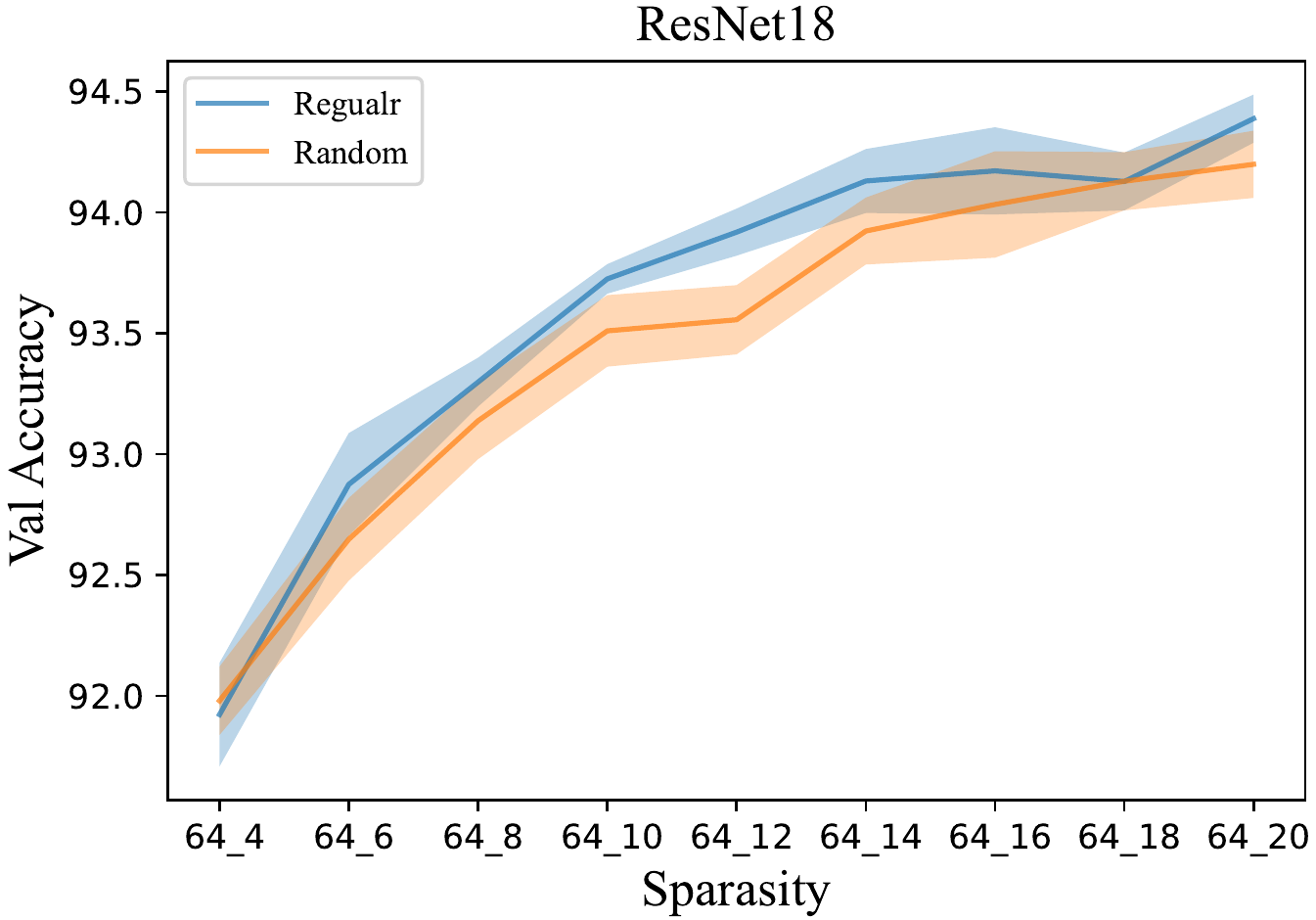}
\caption{Comparison of regular graph based pruning and random graph based pruning (ResNet18 with Cifar10).}
\label{Fig:vs10}
\end{figure}

\begin{figure} [!t]
\includegraphics[width=0.49\textwidth]{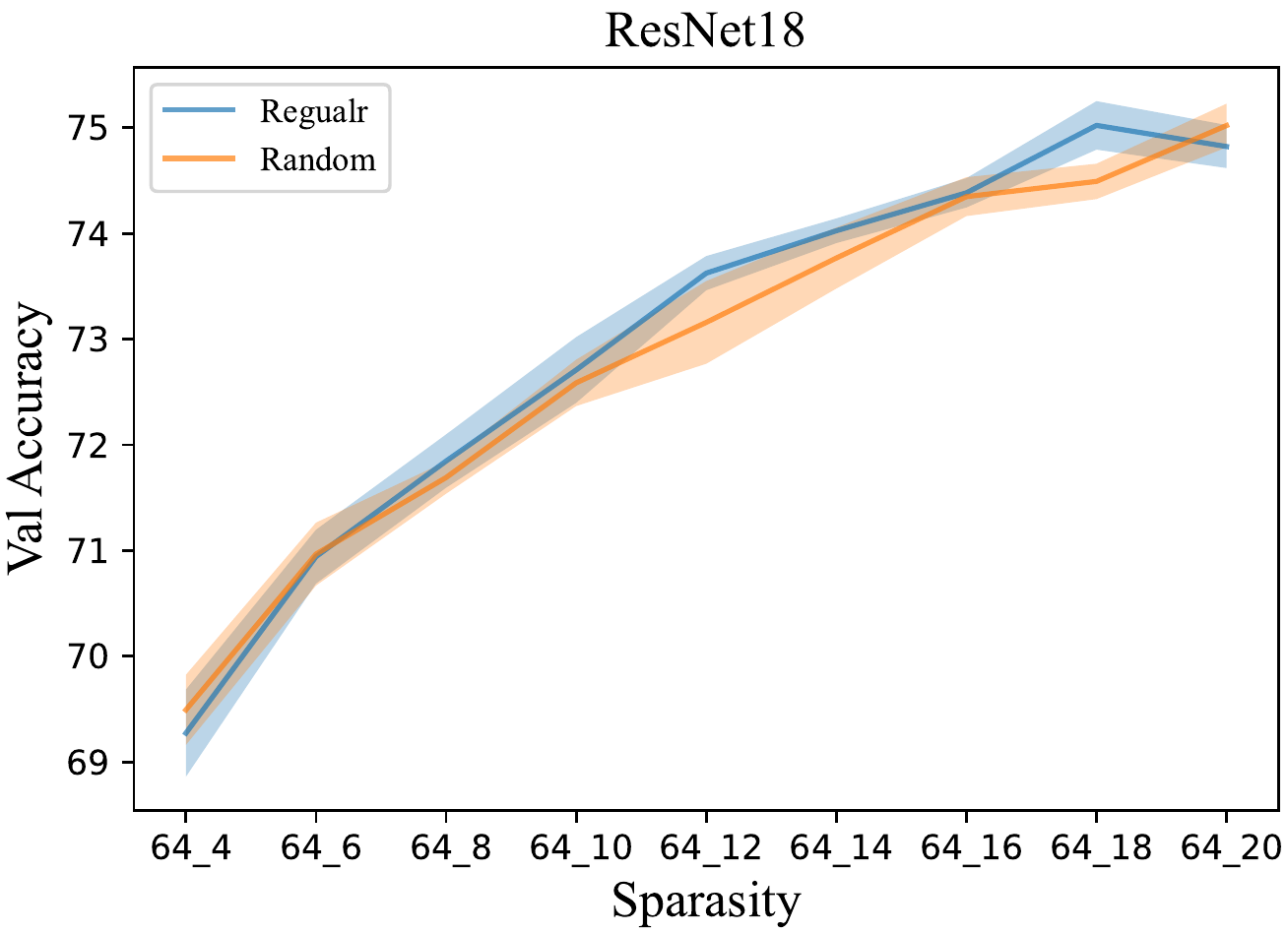}
\caption{Comparison of regular graph based pruning and random graph based pruning (ResNet18 with Cifar100).}
\label{Fig:vs100}
\end{figure}

\subsection{Comparison of regular graph based pruning and random graph based pruning}
We test the performance of random graph based pruning through experiments, and compare it with the performance of regular graph based pruning under the same sparsity, as shown in Figure~\ref{Fig:vs10} and Figure~\ref{Fig:vs100}. As we can see, for ResNet18 with Cifar10/Cifar100, in most sparsity, the performance of regular graph based pruning is better than random graph based pruning, which shows that choosing regular graph as the optimization space is a more suitable choice.

\subsection{Additional experiment on ImageNet}
We also evaluate the performance of RGP on the large-scale dataset ImageNet~\cite{deng2009imagenet} and compare it with other methods. 

We use ResNet50~\cite{he2016deep} as the original model, set the batch size to 256 and the weight decay to 0.0001. The initial learning rate is 0.1 and multiplies 0.1 at epoch 30, 60, and 90 with a total of 100 training epochs, which is the same as the training strategy of~\cite{wang2020accelerate}.

\begin{table*}
  \caption{Comparison results with other methods on ImageNet, the pruned model is ResNet50.}
  \label{table3}
  \centering
  \resizebox{0.7\linewidth}{!}{\large
  \begin{tabular}{cccccc}
    \toprule
     Method & \makecell[c]{Original\\Top1 Acc} & \makecell[c]{Pruned\\Top1 Acc} & Acc Drop & \makecell[c]{Parameters \\Reduction} & \makecell[c]{FLOPs \\Reduction} \\
    \midrule
    GAL~\cite{lin2019towards} & 76.15\% & 71.95\% & 4.20\% & 16.90\% & 43.00\% \\
    ThiNet~\cite{luo2017thinet} & 72.88\% & 72.04\% & 0.84\% & 33.72\% & 36.80\% \\
    SFP~\cite{DBLP:conf/ijcai/HeKDFY18} & 76.15\% & 74.61\% & 1.54\% & - & 41.80\% \\
    Taylor~\cite{molchanov2016pruning} & 76.18\% & 74.50\% & 1.68\% & 44.50\% & 44.90\% \\
    Autopruner~\cite{luo2020autopruner} & 76.15\% & 74.76\% & 1.39\% & - & 48.70\% \\
    C-SGD~\cite{ding2019centripetal} & 75.33\% & 74.93\% & 0.40\% & - & 46.20\% \\
    HRank-1~\cite{lin2020hrank} & 76.15\% & 74.98\% & 1.17\% & 36.70\% & 43.70\% \\
    \textbf{RGP (64\_36)} & 76.22\% & 75.30\% & 0.92\% & 43.75\% & 43.75\% \\
    PFP-A~\cite{Liebenwein2020Provable} & 76.13\% & 75.91\% & 0.22\% & 18.10\% & 10.80\% \\
    PFP-B~\cite{Liebenwein2020Provable} & 76.13\% & 75.21\% & 0.92\% & 30.10\% & 44.00\% \\
    HRank-2~\cite{lin2020hrank} & 76.15\% & 71.98\% & 4.17\% & 62.10\% & 46.00\% \\
    RRBP~\cite{zhou2019accelerate} & 76.10\% & 73.00\% & 3.10\% & - & 54.50\% \\
    HRank-3~\cite{lin2020hrank} & 76.15\% & 69.10\% & 7.05\% & 76.04\% & 67.57\% \\
    \textbf{RGP (64\_30)} & 76.22\% & 74.58\% & 1.64\% & 53.13\% & 53.13\% \\
    \textbf{RGP (64\_24)} & 76.22\% & 74.07\% & 2.15\% & 62.50\% & 62.50\% \\
    Prune-From-Scratch~\cite{wang2020pruning} & 77.20\% & 72.80\% & 4.40\% & - & 75.61\% \\
    \textbf{RGP (64\_16)} & 76.22\% & 72.68\% & 3.54\% & 75.00\% & 75.00\% \\
    \bottomrule
  \end{tabular} }
\end{table*}

We chose more pruning methods for comparison, and the results are shown in Table~\ref{table3}. We can see that when the parameters reduction and the FLOPs reduction are both within 50\%, our method (RGP(64\_36)) is better than GAL~\cite{lin2019towards}, ThiNet~\cite{luo2017thinet}, SFP~\cite{DBLP:conf/ijcai/HeKDFY18}, HRank-1~\cite{lin2020hrank} and PFP-B~\cite{Liebenwein2020Provable} in terms of both pruning ratio and top1 accuracy, and is better than Autopruner~\cite{luo2020autopruner} and C-SGD~\cite{ding2019centripetal} in terms of top1 accuracy at a close pruning ratio. PFP-A~\cite{Liebenwein2020Provable} has the highest top1 accuracy, but 
its pruning ratio is also extremely low. At a pruning ratio greater than 60\%, our method (RGP (64\_24)) is better than HRank-3~\cite{lin2020hrank} in top1 accuracy, and at a pruning ratio greater than 70\%, our method (RGP (64\_16)) is close to Prune-From-Scratch~\cite{wang2020pruning} in top1 accuracy. In general, our method shows a great pruning effect, and can greatly reduce the amount of neural network parameters and FLOPs. Compared with other pruning methods, our method has a simpler process, the obtained graph can be used in a large number of neural network structures, and the pruning process does not require the dataset to participate in parameter training, which is an efficient and one-shot method.

\clearpage

\end{document}